\documentclass{ifacconf}

\usepackage{graphicx}      
\usepackage{natbib}        
\usepackage{amsmath,amssymb,amsfonts}
\usepackage{url}
\usepackage{nomencl}
\usepackage{xpatch}
\usepackage{tabularx}
\usepackage{multirow}
\usepackage{multicol}
\usepackage{svg}
\usepackage{graphics} 
\usepackage{newunicodechar}
\usepackage{siunitx}
\usepackage{diagbox}
\usepackage{xcolor}
\usepackage{wrapfig}
\usepackage{mathtools}
\usepackage{array}
\usepackage{balance}

\begin{document}
\begin{frontmatter}

\title{Fast Neural-Network Approximation of Active Target Search Under Uncertainty}

\author{Bilal Yousuf$^{*}$, Zs\'{o}fia Lendek$^{*}$, Lucian Bu\c{s}oniu$^{*+}$}
\address{$^{*}$Technical University of Cluj-Napoca, Romania\\
$^{+}$Corresponding Member of Romanian Academy \\ e-mail: \{bilal.yousuf, zsofia.lendek, lucian.busoniu\}@aut.utcluj.ro}

\begin{abstract}                
We address the problem of searching for an unknown number of stationary targets at unknown positions with a mobile agent. A probability hypothesis density filter is used to estimate the expected number of targets under measurement uncertainty. Existing planners, such as Active Search (AS) and its Intermittent variant (ASI), achieve accurate detection but require costly online optimization. To reduce online computation, we propose to use a convolutional neural network to approximate AS or ASI decisions through direct inference. The network is trained on AS/ASI data using a multi-channel grid that encodes target beliefs, the agent position, visitation history, and boundary information. Simulations with uniform and clustered target distributions show that the network achieves detection rates comparable to AS or ASI while reducing computation by orders of magnitude.
\vspace*{-1em}
\end{abstract}

\end{frontmatter}

\section{Introduction}
Searching for multiple targets in an unknown environment is a fundamental problem in robotics, with applications ranging from search-and-rescue \citep{Pallin:2021, Cooper:2020} and environmental monitoring \citep{Khan:2024} to surveillance \citep{Kolios:2021} and exploration \citep{Olcay:2020, Aguilar:2019}. The core challenge lies in designing planners that can efficiently guide autonomous agents to detect the targets while balancing exploration of unvisited areas and refinement of uncertain target estimates. Model-based planners \citep{Dam:2015}, often framed as Model Predictive Control (MPC), address this challenge by repeatedly solving optimization problems online, but this can become computationally expensive. In parallel, data-driven approaches, particularly those based on Neural Networks (NNs), have emerged as powerful alternatives for approximating such decision-making processes \citep{Ren:2024}. 
By learning a direct mapping from input features to actions, NN-based methods can implicitly capture the predictive behavior of MPC while drastically reducing computation time, leading to an explicit MPC approach \citep{Bem09}.

Recent research has explored NNs and deep-learning methods for target search, exploration, and navigation. CNN-based architectures have been successfully used for vision-driven navigation \citep{Zhu:2017, Yuanda:2018}, autonomous exploration \citep{Chen:2019}, and underwater mapping for marine robotics \citep{Zhong:2024}. In the specific context of target search, deep learning has been applied to improve detection and localization performance under uncertainty \citep{Khan:2024, Ge:2024}, showing that data-driven planners can significantly reduce online computational demands. However, these approaches often have difficulty balancing exploration of new areas with the refinement of uncertain target estimates \citep{Kolios:2021, Lixin:2019}, and they are not well integrated with probabilistic filtering frameworks designed to handle highly uncertain sensor information.

To address this gap, we propose a Convolutional Neural Network (CNN)-based adaptation of Active target Search (AS) \citep{Bilal:24} as well as of AS with Intermittent measurements (ASI) \citep{Bilal:25}. A Probability Hypothesis Density (PHD) filter propagates an intensity function whose integral over any region yields the expected number of targets in that region. AS selects waypoints by optimizing a target-refinement and exploration objective at each step, while ASI extends this approach to longer trajectories for their waypoints and intermittent measurements. Both algorithms find targets quickly in simulation and real-life experiments, but their reliance on online optimization makes them computationally expensive during execution, which is what our CNN approach aims to improve. The CNN  approximates the decision-making process of AS and ASI by taking as input a multi-channel spatial grid representation that encodes particle filter outputs, visitation counts, boundary masks, and the agent position, and directly predicts a near-optimal next waypoint. By doing so, the CNN inherits the strengths of AS and ASI while drastically reducing computational cost. We also introduce new exploration-free versions of AS and ASI, motivated by the fact that explicit exploration terms can sometimes drive agents away from regions of actual target likelihood, which may reduce efficiency in sparse-target environments. The search for new targets is instead naturally guided by the birth intensity in the PHD filter.

We validate our framework through extensive simulation experiments for both uniform and clustered target distributions. The experiments include evaluations of target discovery performance against original AS and ASI, computational efficiency as a function of the number of candidate waypoints, and ablation studies of the CNN input channels. Across all scenarios, our results show that the CNN achieves detection rates that are statistically indistinguishable from the original methods, while being orders of magnitude faster.

Next, Section \ref{sec:problem} formulates the problem, followed by an overview of PHD filtering and model-based target search in Section \ref{sec:filter}. The proposed CNN-based method is then presented separately in Section \ref{sec:methods}. Simulation experiments are reported in Section \ref{sec:simulation}, and Section \ref{sec:conclusion} concludes.

\section{Problem Formulation}\label{sec:problem}
\begin{figure}[h!]
  \centering
  \includegraphics[scale=0.075]{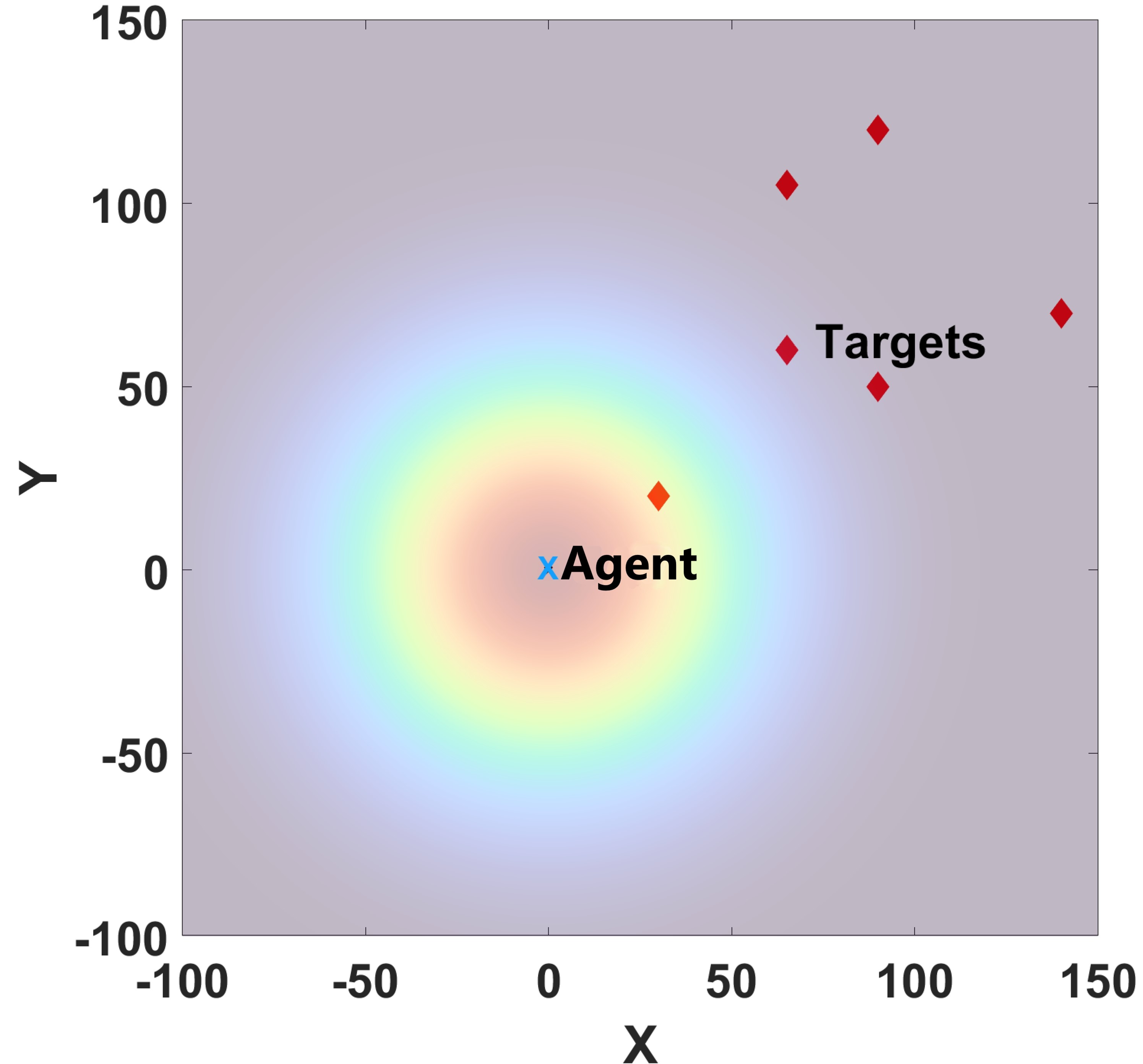}
  \caption{2D space with 6 targets and a single agent with a circular field of view. The dark orange, yellow, and blue colors show the probability of target observation (higher to lower) from the current position of the agent.}
  \label{Fig:Prob_of_Detection}
\end{figure}

Consider an agent navigating a 2D target space (environment) $E$ in search of an unknown number of static targets, as depicted in Fig.~\ref{Fig:Prob_of_Detection}. The main objective is to find the positions of all the targets in as few steps as possible. 


The agent dynamics are given by:
\begin{equation} \label{eq:drone_dynamics}
q_{k+1}=f(q_{k},u_{k})
\end{equation}
with $k$ being the discrete time step. The state $q_{k}\thinspace \in \thinspace \Re^{n_{q}}$, where $n_{q}$ is the dimension of the state of the agent, contains the position $\tilde{q}_{k}=[\mathtt{X}_{k}, \mathtt{Y}_{k}]$ and will typically also include orientation, linear, and angular velocities. The input is $u_{k} \thinspace \in \thinspace \Re^{n_{u}}$. The control law can be e.g.,~a state feedback: $u_{k}=h(q_{k})$, but other controllers may be used. 

Environment $E$ contains $N$ stationary targets, and each target $i$ is located at coordinates $x_{i}=(\mathtt{X}_{i},\mathtt{Y}_{i})\thinspace\in\thinspace E$, $i=1,2,\hdots,\thinspace N$. The set of targets is denoted by $X$. Both the cardinality $N$ and the locations of the targets are initially unknown. We use a random finite set framework \citep{Dam:2015, Dou:2005} to estimate target locations, as detailed in \citep{Bilal:25}. At each step, agents receive noisy measurements about a subset of targets that are detected. The probability with which the agent at coordinates $\tilde{q}=[\mathtt{X},\mathtt{Y}]$ detects a target at position $x_{i}=[\mathtt{X}_{i}, \mathtt{Y}_{i}]$ is denoted by $\pi(x_{i}, \tilde{q})$, see again Fig. \ref{Fig:Prob_of_Detection}. In all simulations, we consider an omnidirectional ranging sensor for which 
\begin{equation}\label{eq:pi}
\pi(x_{i},\tilde{q})= G e^{-\left\|\zeta\right\|/2}
\end{equation}
where scalar $G \leq 1$, and $\zeta = \left(\frac{\mathtt{X}_{i}-\mathtt{X}}{\mathbb{F}_{\mathtt{X}}},\frac{\mathtt{Y}_{i}-\mathtt{Y}}{\mathbb{F}_{\mathtt{Y}}}\right)$ is a normalized distance between the target $x_{i}$ and the agent. Normalization constants $(\mathbb{F}_{\mathtt{X}},\mathbb{F}_{\mathtt{Y}})$ may be interpreted as the size of the probabilistic field of view of the agent.  For example, when $\mathbb{F}_{\mathtt{X}}=\mathbb{F}_{\mathtt{Y}}$, $\pi$ is radially symmetric around the agent position, as illustrated in Fig~\ref{Fig:Prob_of_Detection}.

Define $k_{m}$ as the time step at which the agent takes the $m$th measurement, where $m\geq0$. Measurements are taken at a multiple $\Delta$ of the control sampling period $T_{s}$. For example, $\Delta=2$ corresponds to $k_{m}=0,2,4,6\hdots\thinspace \thinspace$. The binary event $b_{ik_{m}}$ of the agent with position $\tilde{q}_{k_{m}}=(\mathtt{X}_{k_{m}},\mathtt{Y}_{k_{m}})$ detecting at step $k_{m}$ a target at position $x_{i}$ follows a Bernoulli distribution given by the probability of detection: $b_{ik_{m}} \sim \mathcal{B}(\pi(x_{i},\tilde{q}_{k_{m}}))$. Given these Bernoulli variables $b_{ik_{m}}$, the set of measurements $Z_{k_{m}}$ is:
\begin{equation}\label{eq:sensor_model}
Z_{k_{m}}=\bigcup_{i\thinspace\in\left\{1,\hdots,N\right\}\thinspace \mathrm{s.t}. \thinspace b_{ik_{m}}=1}z_{k_{m}}
\end{equation}
where $z_{k_{m}}=g_{k_{m}}(x_{i})+\varrho_{k_{m}}$, and $g_{k_{m}}(x_{i})$ is defined as:
\begin{equation*}
\begin{aligned}
&g_{k_{m}}(x_{i})=\left[r_{ik_{m}}, \theta_{ik{m}}\right]^{T}\\
&r_{ik_{m}}= \sqrt{(\mathtt{X}_{i}-\mathtt{X}_{k_{m}})^2+(\mathtt{Y}_{i}-\mathtt{Y}_{k_{m}})^{2}}\\
&\theta_{ik_{m}}= \arctan\frac{\mathtt{Y}_{i}-\mathtt{Y}_{k_{m}}}{\mathtt{X}_{i}-\mathtt{X}_{k_{m}}}
\end{aligned}
\end{equation*}
So, for each target $i$ detected, the range $r_{ik_{m}}$ and bearing angle~$\theta_{ik_{m}}$ relative to the agent are measured. This measurement is affected by Gaussian noise $\varrho_{k_{m}} \sim\mathcal{N}(,\mathbf{0},R)$, with mean $\mathbf{0} = [0,0]^\top$~and diagonal covariance $R=\mathrm{diag}[(\sigma^{a})^{2}, (\sigma^{a})^{2}]$. Thus, the target measurement density $p(z_{k_{m}}|x)$ is a Gaussian density centered on $g_{k_{m}}(x_{i})$ with covariance matrix $R$.

\section{Preliminaries} \label{sec:filter}
This section summarizes the key preliminaries required for the proposed methods. We begin with the PHD filter, adapted from \citet{Dou:2005}, in Section \ref{subsec:PHD}. Section \ref{AS} then introduces the Active target Search (AS) approach, followed by Active target Search with Intermittent measurements (ASI) in Section \ref{ASI}. Both methods generate waypoints to guide the agent through the environment, but they differ in their optimization objectives and how they take measurements.

\subsection{Probability Hypothesis Density Filter} \label{subsec:PHD}
Define first the intensity function $I:E\rightarrow[0,\infty)$, which is similar to a probability density function, with the key difference that its integral $\int_{S} I(x) dx$ over some subset $S \subseteq E$ is not the probability mass of $S$, but the expected number of targets in $S$.

The PHD filter includes a prediction step $\Phi$ that propagates the intensity function, and an update step $\Psi_{k_{m}}$, that performs Bayesian updates of an intensity function based on the measurements $Z_{k_{m}}$:
\begin{equation}\label{eq:summerized}
\begin{aligned}
&I_{k|k-1}=\Phi(I_{k-1|k-1})\\
&I_{k|k}=
\begin{cases}
\Psi_{k_{m}}(I_{k_{m}|k_{m}-1},Z_{k_{m}}), & \text{ if } k=k_{m} \mbox{ for } m\geq 0\\ 
I_{k|k-1},& \text{ otherwise }  
\end{cases}
\end{aligned}
\end{equation}
Here, $I_{k|k-1}$ is the prior intensity function, predicted based on $I_{k-1|k-1}$ at the previous step, and $I_{k_{m}|k_{m}}$ denotes the posterior generated after processing the new measurements at $k_{m}$. By convention, at $k\neq k_{m}$, $I_{k|k}=I_{k|k-1}$. The prediction step is defined as:
\begin{equation}\label{eq:PHD_predict}
\begin{aligned}
\Phi(I_{k-1|k-1})(x)= \Upsilon+\int_{E} p_{s}(\xi)\delta_{\xi}(x) I_{k-1|k-1}(\xi) d\xi
\end{aligned}
\end{equation}
where $p_{s}(\xi)$ is the probability that a target previously located at position $\xi$ still exists. Targets are stationary, so the transition density of a target $x$ at $\xi$ is defined as the Dirac delta $\delta_{\xi}(x)$ centered on $\xi$. Moreover, $\Upsilon$ denotes the intensity function of a new target appearing, which is taken here constant across the environment because we do not assume any prior information on where new targets will appear. The update at step $k_{m}$ is: 
\begin{equation}\label{eq:PHD_update}
\begin{aligned}
&\Psi_{k_{m}}(I_{k_{m}|k_{m}-1}, Z_{k_{m}})(x)=\\
&{\left[1-\pi(x_{i},\tilde{q}_{k_{m}})+\sum_{z\thinspace\in\thinspace Z_{k_{m}}}\frac{\psi_{k_{m}z}(x) }{\left<\psi_{k_{m}z},I_{k_{m}|k_{m}-1}\right>}\right]}\cdot I_{k_{m}|k_{m}-1}(x)
\end{aligned}
\end{equation}
where $\psi_{k_{m}z}(x)=\pi(x_{i},\tilde{q}_{k_{m}})p(z|x)$ denotes the overall probability density of detecting a target at $x$, via a measurement $z$ with $p$, and ~$\left<\psi_{k_{m}z},I_{k_{m}|k_{m-1}}\right>=\int_{E}\psi_{k_{m}z}(x)I_{k_{m}|k_{m-1}}(x)dx$. In practice, we apply the sequential Monte-Carlo PHD filter \citep{Dou:2005}, which uses at each $k$ a set of weighted particles $(x^{j}, \varpi_{k|k}^{j})$ to represent $I_{k|k}$, with the property that $\int_{S} I_{k|k} (x) dx \approx \sum_{x^{j} \in S} \varpi_{k|k}^{j}$ for any $S \subseteq E$. For details, see \cite{Bilal:25}.

\subsection{Active Target Search (AS)} \label{AS}
The AS planner, introduced in \citep{Bilal:24}, generates a sequence of waypoints for the agent by evaluating at step $k$ a discrete set of candidate waypoints  $\mathcal{W}_{k}=\left\{w^{o}_{k},\thinspace\thinspace o=1,\hdots,W \right\}$ generated around $\tilde{q}_{k}$, where $w_{k}^{o}$ is the $o$th candidate waypoint, and $W$ is the number of candidate waypoints. In this setting, measurements are taken at each step, so $\Delta=1$  which corresponds to $k_{m}=k=0,1,2,3\hdots\thinspace \thinspace$. At each step $k$, the next waypoint is obtained by solving the optimization problem:
\begin{equation}\label{eq:optimizationn}
{o}_{k}^{*}\in\mathrm{argmax}_{o=1,\hdots,W} \mathbb{T}_{k}(o) 
\end{equation}
where candidate targets are extracted as clusters of particles using K-means clustering \citep{Dou:2005}, and the refinement objective $\mathbb{T}_{k}(o)$ is defined as
\begin{equation}\label{eq:co}
\mathbb{T}_{k}(o) =\sum_{c=1}^{C_k} \pi(\nu_{k}^{c},\tilde{q}_{k})
\end{equation}
where $C_k$ denotes the number of clusters at $k$, and $\nu_{k}^{c}$ is the center of the $c$th cluster, interpreted as an estimated target position.

In our earlier work, we used an explicit exploration term in AS to encourage agents to explore the environment and locate new targets. In this paper, we instead introduce an exploration-free version of AS, where discovery of new targets is guided by the birth intensity $\Upsilon$ of the PHD filter in \eqref{eq:PHD_predict}. This mechanism naturally introduces new particles across the environment, with weights that grow over time, gradually promoting new target discovery via altering $\mathbb{T}_{k}(o)$ in \eqref{eq:co}.

\subsection{Active Target Search with Intermittent Measurements}\label{ASI}
ASI \citep{Bilal:25} extends AS by incorporating control costs and Intermittent measurement updates. At each decision index $d$, that occurs at the discrete time step $k_{d}$,  the agent evaluates trajectories from its current state $q_{k_{d}}$ to candidate waypoints $w_{d}^{o}$. Each waypoint $w_{d}^{o}$ is passed to a low-level controller, which generates a~predicted trajectory~$\mathbf{q}_{d}^{o}=[q_{k_{d}+1}^{o},\ldots,q_{k_{d}+K_{d}^{o}}^{o}]$ and corresponding control inputs $\mathbf{u}_{d}^{o}=[u_{k_{d}}^{o},\ldots,u_{k_{d}+K_{d}^{o}-1}^{o}]$, where $K_{d}^{o}$ is the number of steps required to reach $w_{d}^{o}$ from the current state $q_{k_{d}}$. Along each trajectory, measurements are taken periodically at $k_{d,\tilde{m}}=k_{d}+\tilde{m}\Delta$ for $\tilde{m}=0,\ldots,M_{d}^{o}-1$, where $M_{d}^{o}=\lceil K_{d}^{o}/\Delta \rceil$ defines the measurement horizon. The optimal waypoint is selected by solving:
\begin{equation}\label{eq:optimization}
o_{d}^{*}\in\mathrm{argmax}_{o=1,\hdots,W}\frac{-\mathbb{C}_{d}(o)+\beta\cdot\mathbb{T}_{d}(o)}{K_{d}^{o}}
\end{equation}
The control cost (effort) to reach the $o$th candidate waypoint is: $\mathbb{C}_{d}(o)=\sum_{j=0}^{K_{d}^{o}-1}(u_{k_{d}+j}^{o})^{T} u_{k_{d}+j}^{o}$, and the refinement component $\mathbb{T}_{d}(o)$ extends the AS objective by accounting for measurements every $\Delta$ steps along the entire predicted trajectory: $\mathbb{T}_{d}(o)=\sum_{\tilde{m}=0}^{M_{d}^{o}-1}\sum_{c=1}^{C_{k_{d}}}\pi(\nu_{k_{d}}^{c},\tilde{q}_{k_{d,\tilde{m}}}^{w})$.\\
The parameter $\beta$ tunes the tradeoff between control cost and target refinement component.

Like in AS, the explicit exploration term used by \cite{Bilal:25} is implicitly replaced by the probabilistic birth process $\Upsilon$. 

For both AS and ASI, to avoid wasting effort on already well-determined targets, narrow particle clusters that accumulated large weights are marked as found targets. Measurements near these confirmed targets are then discarded to prevent re-detection; for details, see \cite{Bilal:24}.

\section{CNN-based Target Search Policies} \label{sec:methods} 
To reduce the computational cost of the AS planner, we introduce a CNN that learns to approximate the optimal waypoint selection process. Training data is generated during the execution of AS and ASI along example trajectories, by recording the agent’s current position $(\mathtt{X},\mathtt{Y})$ together with the set of particles of the filter at each decision step. A 4-channel spatial grid serves as input to the CNN: $\mathcal{I}_{\text{CNN}} \in \mathbb{R}^{n_{g}\times n_{g}\times 4}$, where $n_{g}$ denotes the number of points along each dimension of the grid. For simplicity, a square grid is considered, but the formulation can be readily generalized to a rectangular grid. The input is paired with a target label $\mathcal{T_{\text{CNN}}}$ representing the optimal waypoint selected by AS or ASI, forming an input–label pair for supervised training.  The training dataset is generated from multiple independent simulation trials, each executed for a fixed number of decision steps, resulting in a total collection of training samples of the formulation $(\mathcal{I}_{\text{CNN}},\mathcal{T_{\text{CNN}}})$.

The first channel among the four encodes the visitation history of the environment; here, locations visited more by the agents receive lower values. A linearly decaying memory of visits is used, with grid cell value $\mathcal{C}_{\mathrm{visit}}(a,b)$ computed as:

\begin{equation*}
    \mathcal{C}_{\mathrm{visit}}(a,b)=\frac{1}{1+\eta(a,b)}
\end{equation*}
where $\eta(a,b)$ counts how many times the agent visited the grid cell, and $a,b \in \{1,.\hdots, n_{g}\}$ are integer cell coordinates on the grid. A visit is registered only when the agent enters a new cell; if the agent remains within the same cell for multiple steps $k$, this is still counted as a single visit.

The second channel represents the agent’s belief on target presence, based on the intensity function. The set of weighted particles is first transformed into a grid-based intensity map, where each cell accumulates the weights of particles that fall within it:
\begin{equation*}
\mathtt{I}(a,b) = \sum_{j:\, x^{(j)} \in \mathtt{c}(a,b)} \varpi^{(j)}
\end{equation*}
where $x^{(j)}$ and $\varpi^{(j)}$ denote the position and weight of the $j$th particle, respectively, and $\mathtt{c} (a,b)$ is the cell at indices $(a,b)$ in the $2$D discretized map. To obtain a smoother representation of belief, the actual channel is obtained by convolving the intensity map with a Gaussian kernel $G_{\sigma}$:
\begin{equation*}
\mathcal{C}_{\mathrm{dens}}(a,b) = (G_{\sigma} * \mathtt{I})(a,b)
\end{equation*}
where $G_{\sigma}(\mathtt{u},\mathtt{v}) = \frac{1}{2\pi\sigma^{2}}\exp\left(-\frac{\mathtt{u}^{2}+\mathtt{v}^{2}}{2\sigma^{2}}\right)$, $*$ denotes $2$D convolution, $\sigma$ is the standard deviation, and $\mathtt{u}, \mathtt{v}$ denote the vertical and horizontal offsets from the center of cell $(a,b)$.

The third channel is a one-hot encoding of the agent's current location:
\begin{equation*}
  \mathcal{C}_{\mathrm{pos}} (a,b)= \begin{cases}
1& \text{if}\;\tilde{q} \in \mathtt{c} (a,b)\\
0 & \text{otherwise}  
\end{cases}
\end{equation*}

Finally, the fourth channel is a boundary-proximity mask that informs the agent about regions that are close to the domain edges, where part of the sensor field of view would fall outside the map and provide less useful information: 
\begin{equation*}
\mathcal{C}_{\mathrm{bound}}(a,b) =
\begin{cases}
1, & \text{if } a \le D \text{ or } a > n_g - D \\
   & \quad \text{or } b \le D \text{ or } b > n_g - D,\\
0, & \text{otherwise.}
\end{cases}
\end{equation*}
where $D > 0$ is an integer defining the boundary thickness in number of grid cells.

The optimal waypoint selected by the AS or ASI model-based planners at each decision step --- using the procedure explained in Sections \ref{AS}-\ref{ASI} --- is represented by coordinates $\bar{\mathtt{X}},\bar{\mathtt{Y}}$, normalized to the unit domain $[0,1]^{2}$, and used as the ground-truth label $\mathcal{T_{\text{CNN}}}$ for supervised training. Later on, when using the CNN for control, its output is rescaled to match the physical size of the environment will be $E$.

\begin{figure}[h!]
\centering
\includegraphics[scale= 0.17]{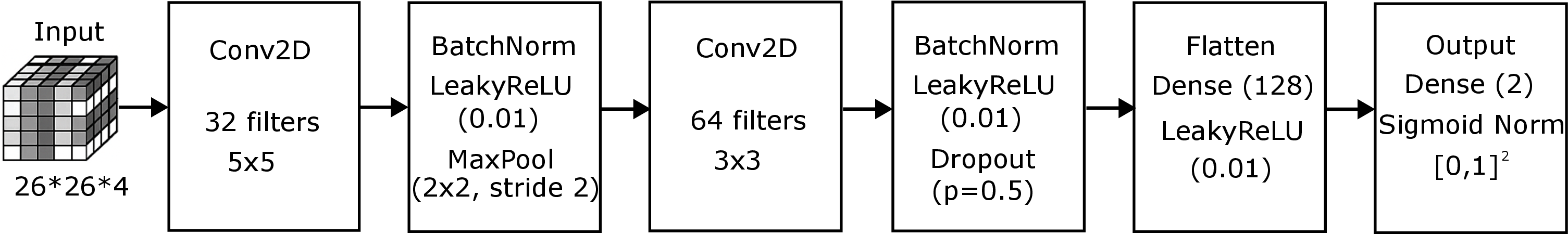}
\caption{Illustration of the CNN structure used for waypoint prediction, where $n_{g}$ is set to 26 like in our experiments, to make the structure definite.}
\label{neural}
\end{figure}

The CNN architecture is, illustrated in Fig~\ref{neural} and consists of an input layer with dimensions $n_{g}\times n_{g}\times4$, followed by two convolutional layers. The first convolutional layer uses $32$ filters of size $5\times5$, followed by batch normalization and Leaky ReLU activation with a slope of $0.01$. A max-pooling layer with stride $2$ reduces spatial resolution. The second convolutional layer applies $64$ filters of size  $3\times 3$, again followed by batch normalization, Leaky ReLU with a slope of $0.01$, and a dropout layer with a probability of $0.5$ to prevent overfitting. The output of the Convolutional layers is passed through two fully connected (dense) layers. The first dense layer has $128$ units with Leaky ReLU activation (slope = $0.01$). The final dense layer outputs a $2$D vector representing normalized spatial coordinates within the unit square $[0,1]^{2}$. A sigmoid activation is used to ensure bounded output. 

The network is trained to minimize the Mean Squared Error (MSE) between the predicted waypoint and the ground truth:

\begin{equation}\label{opt}
L = \frac{1}{\mathcal{S}} \sum_{s=1}^{\mathcal{S}} \left\| \hat{\mathcal{T}}_{\text{CNN}}^{s} - \mathcal{T}_{\text{CNN}}^{s} \right\|^2
\end{equation}
where $\mathcal{S}$ is the total number of samples, and $\hat{\mathcal{T}}_{\text{CNN}}^{s}$ is the output predicted by the neural network for samples. The optimizer used is Adam with adaptive learning rate scheduling. Hyperparameter details are provided in the appendix section in the paper.

In the AS case, the predicted waypoints are smoothed using exponential averaging:
\begin{equation}\label{smooth}
w_{k}=\alpha w_{k-1}+(1-\alpha)\hat{\mathcal{T}}_{\text{CNN},k}
\end{equation}
where $\alpha \in [0,1)$ controls the amount of smoothing, and $w_{k-1}$ is the previously executed waypoint. Smaller values of $\alpha$ make the motion more reactive but less smooth, whereas larger values produce smoother trajectories at the cost of slower adaptation to newly emerging target beliefs. Although this exponential averaging does not enforce a constant step length, the resulting steps are unlikely to exceed those in the original AS planner. This smoothing is applied only when the CNN is trained on AS data, since AS trajectories were originally designed with shorter step lengths. For ASI, no extra smoothing is required because the CNN directly produces more widely spaced waypoints, consistent with the ASI approach.

Once trained, the CNN replaces the AS or ASI planners. Instead of evaluating all candidate waypoints in the model-based planners, the agent now directly queries the trained CNN model using its current position, intensity function, and the other channels, and obtains a predicted, near-optimal waypoint.

\section{Experiments} \label{sec:simulation}
To validate the effectiveness of our proposed target search algorithm, we conducted four simulation experiments ($E1$–$E4$). In $E1$, we compared our CNN-based planner with Active Search (AS). In $E2$, the CNN was compared to (and trained on) trajectories from Active Search with Intermittent Measurements (ASI). In $E3$, we analyzed the impact of varying the number of candidate waypoints on AS and ASI, compared to the CNN planner. In $E4$, we performed an ablation study of the CNN architecture to assess the contribution of individual input channels by comparing the performance of the full 4-channel CNN with simplified variants obtained by removing or simplifying one channel at a time.

Experiments $E1$–$E2$ were run under two target distributions—uniform and clustered—over 20 randomized trials each. The other two experiments, $E3$ and $E4$, used only the more challenging clustered setup. The uniform distribution placed 18 targets uniformly randomly across the environment. In the clustered distribution, targets were grouped into three well-separated clusters within the environment. Each cluster center was randomly chosen but kept apart from boundaries and other clusters, with five nearby targets placed around it to form compact groups. 

All experiments run in the 2D environment~$E=[0,260]$m$\times$\\
$[0,260]$m. For simplicity, we use a linear model of a Parrot Mambo drone as our agent \citep{suto:2023}, as our objective is to evaluate the performance of the planner, rather than that of the low-level control. The control sampling period is $T_{s}=0.005$s and the measurement period is $\Delta=10$ times $T_{s}$, i.e., $0.05$s.  To ensure a fair comparison, the initial position of the agent was fixed for all experiments and across all methods at $\tilde{q}_0 = [10, 10]^T$. The parameters of the probability of detection $\pi(x_{i},\tilde{q})$ from Section \ref{sec:problem} are: $G=0.98$, $\mathbb{F}_{\mathcal{X}}=\mathbb{F}_{\mathcal{Y}}=25$. The measurement noise covariance matrix in model \eqref{eq:sensor_model} is $R=\mathrm{diag}[1.5, 0.175]$. We set the maximum number of particles to 5000.

The agents executed 250 steps $k$ in all AS experiments, whereas ASI was run for 50 decision indices $d$. These settings are sufficient to find all targets. Recall that, in AS, a measurement is taken at every step, while in ASI, multiple measurements are collected along each longer trajectory. Performance was evaluated by the average number of targets detected, with results reported alongside 95\% confidence intervals over 20 trials.

The CNN model operates on a spatial grid of size $26 \times 26$ with each cell corresponding to a $10\times10$~m$^{2}$ region of the environment. To generate training data, we conducted $60$ independent simulation trials with uniformly distributed targets. For the AS, each trial was executed for $250$ steps $k$, resulting in a total of $9120$ training samples. For ASI, each trial was run for $50$ decision indices $d$, yielding $3,120$ training samples in total. In AS, the smoothing factor $\alpha$ in \eqref{smooth} was set to 0.7 to provide a good compromise between waypoint smoothness and responsiveness. The CNN was trained for $100$ epochs using these datasets, with early stopping and dropout regularization applied to prevent overfitting, as illustrated in Fig. \ref{neural}.

\medskip \noindent \textit{E1: Comparison of CNN to AS.} We begin our evaluation by comparing the proposed CNN-based planner with the original Active Search (AS) method, aiming to detect targets distributed either uniformly or in clusters. To ensure a fair comparison, both methods operate under the same constraints: they are allowed an equal number of steps, and each takes measurements at approximately 9 meters of travel between consecutive sensing actions. This allows us to evaluate whether a data-driven planner can reproduce the performance of AS under similar movement and sensing. At each planning step, the AS agent selects its next position from a discrete set of eight candidate directions, denoted by $\mathcal{W}$. These correspond to movements to the top, bottom, left, and right, as well as the four diagonal directions oriented at $45^\circ$ angles. In contrast, the CNN directly predicts a waypoint anywhere in the continuous environment, and then smooths the prediction with \eqref{smooth}. 

\begin{figure}[h!]
    \centering
    \includegraphics[scale=0.18]{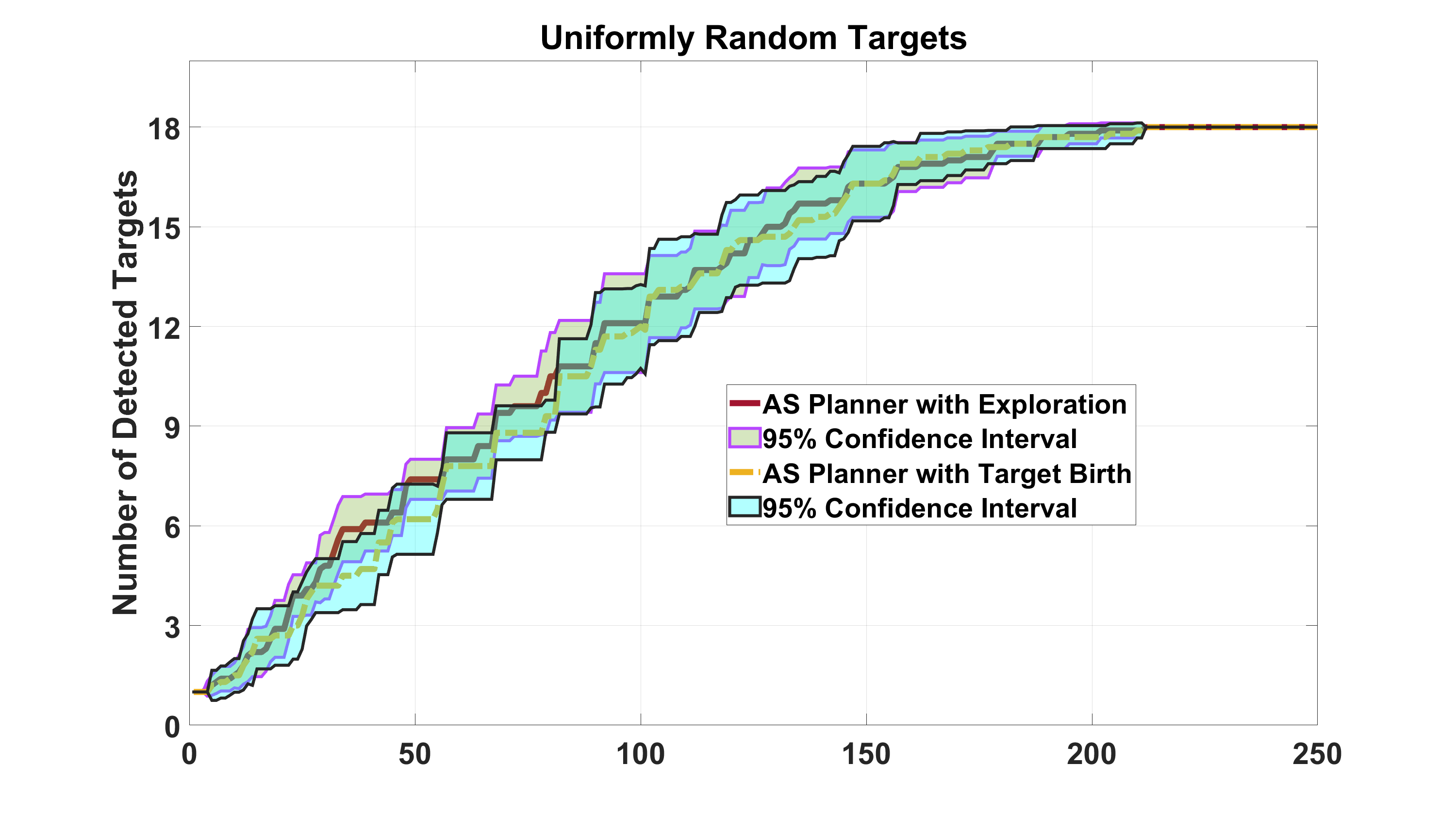}
    \caption{Performance of the AS planner with exploration versus AS with a target birth term}
    \label{Figureexpvstarbi}
\end{figure}

For completeness, we performed a preliminary experiment under the uniform target distribution to assess the effect of removing explicit exploration from the AS planner while introducing a target birth term. The results, shown in Fig.~\ref{Figureexpvstarbi}, indicate performance similar to the standard AS with exploration from \cite{Bilal:24}. This demonstrates that incorporating the target-birth term allows the planner to remain competitive without explicit exploration. In the sequel, we only use the target birth versions of AS and ASI.

\begin{figure}[h!]
\centering
    \includegraphics[scale=0.20]{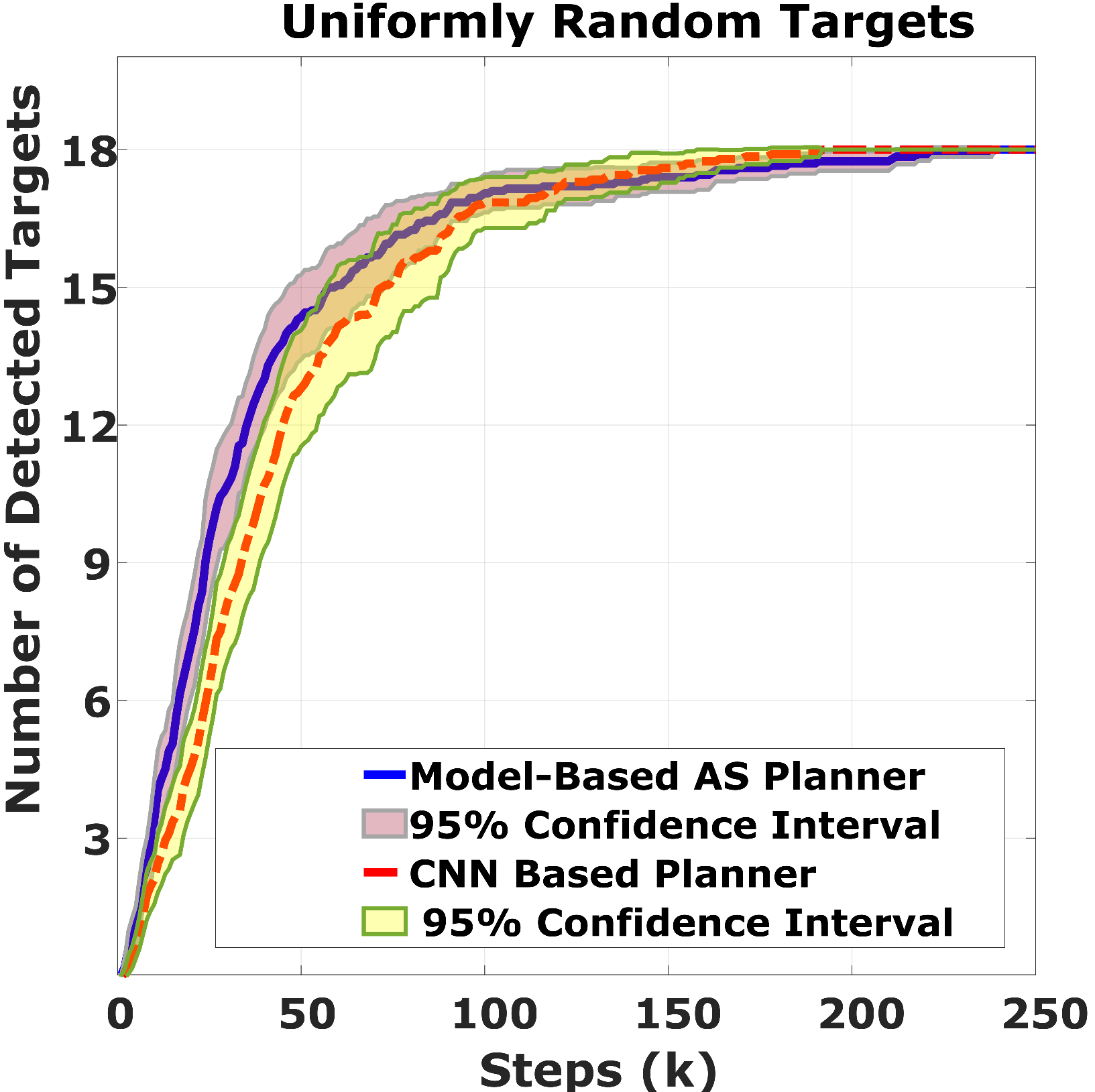}
     \includegraphics[scale=0.3]{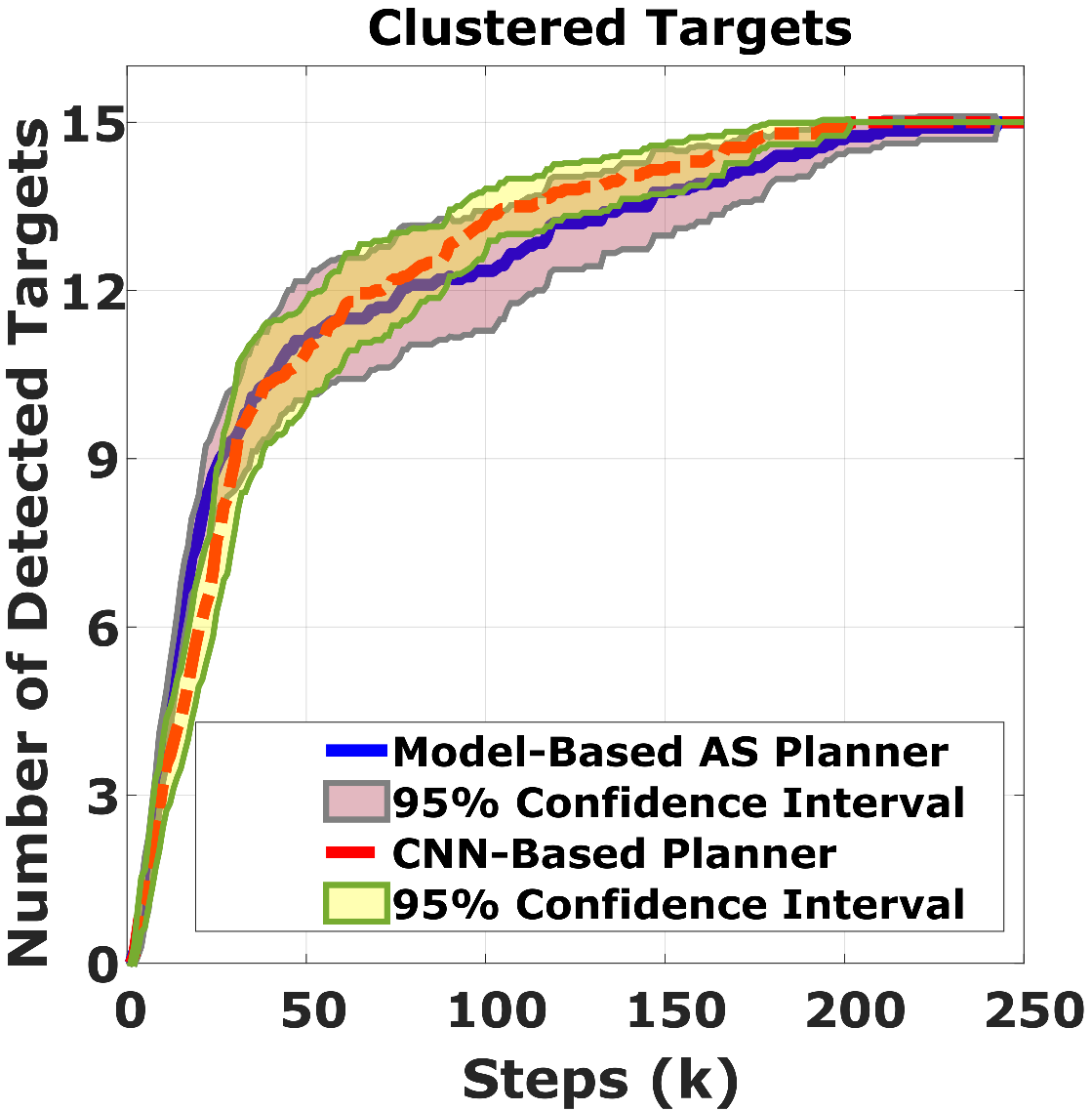}
    \caption{Number of targets detected using the CNN and AS.  Top: Uniformly random targets. Bottom: Clustered targets.}
    \label{Figureplanner}
\end{figure}

The results in Fig.~\ref{Figureplanner} show the number of target detections as a function of the number of steps $k$. The two planners achieve statistically similar performance. However, in the early steps, the CNN planner lags slightly behind AS. In the final phase, CNN catches up and often locates the last few targets faster. Regarding computation time, AS requires $1.4\times 10^{-3}\pm 2.7\times 10^{-3}$s per step, while CNN requires $1.3\times 10^{-4}\pm 9.0\times 10^{-5}$s, averaged across waypoints. This demonstrates that CNN learns a competitive policy with near-identical performance to AS, while providing significantly faster inference.\footnote{The computer used is equipped with an Intel 1365U CPU, 32 GB of RAM, and runs Matlab R2024.}

\begin{figure}[h!]
\centering
    \includegraphics[scale=0.22]{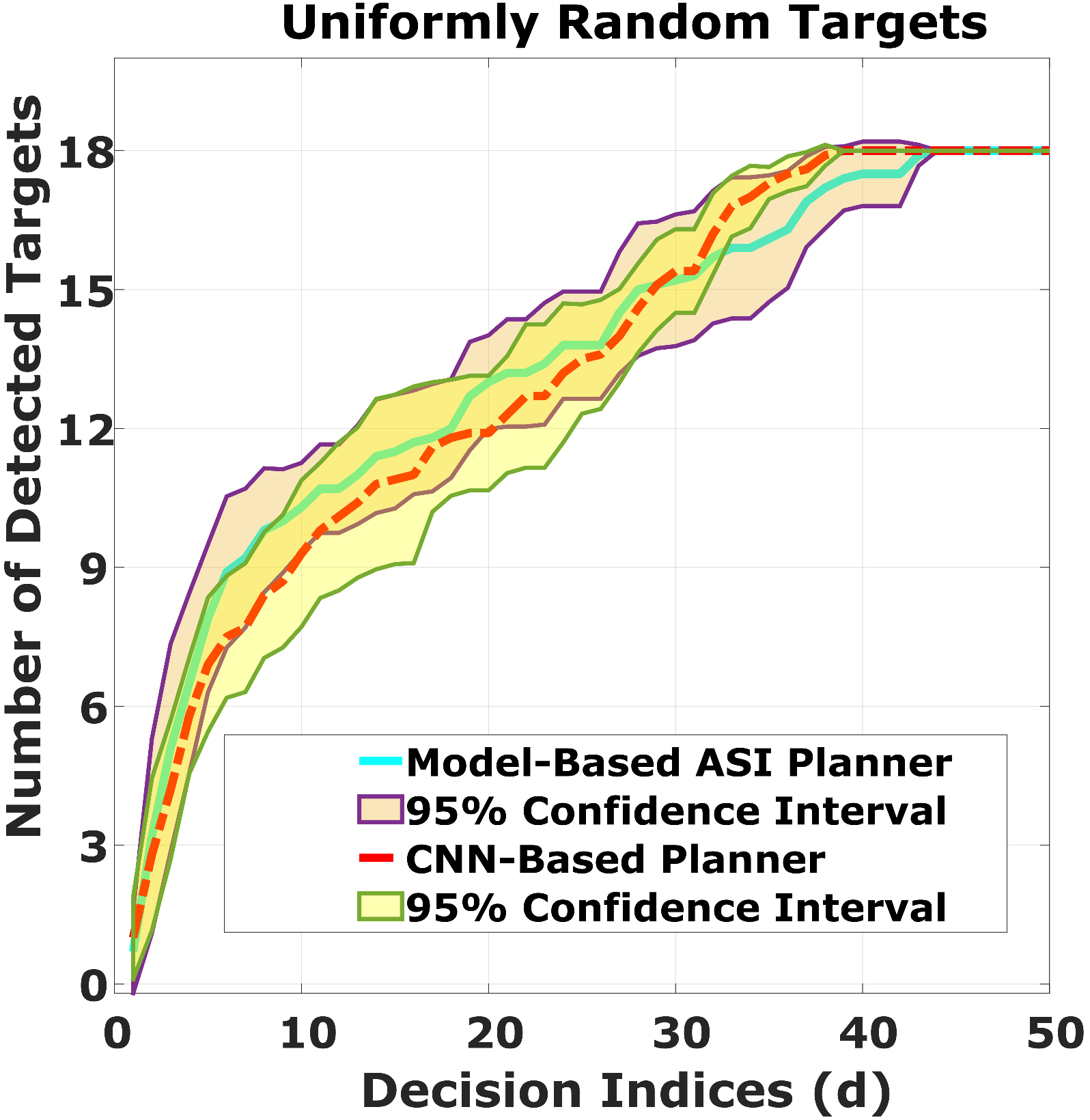}
     \includegraphics[scale=0.22]{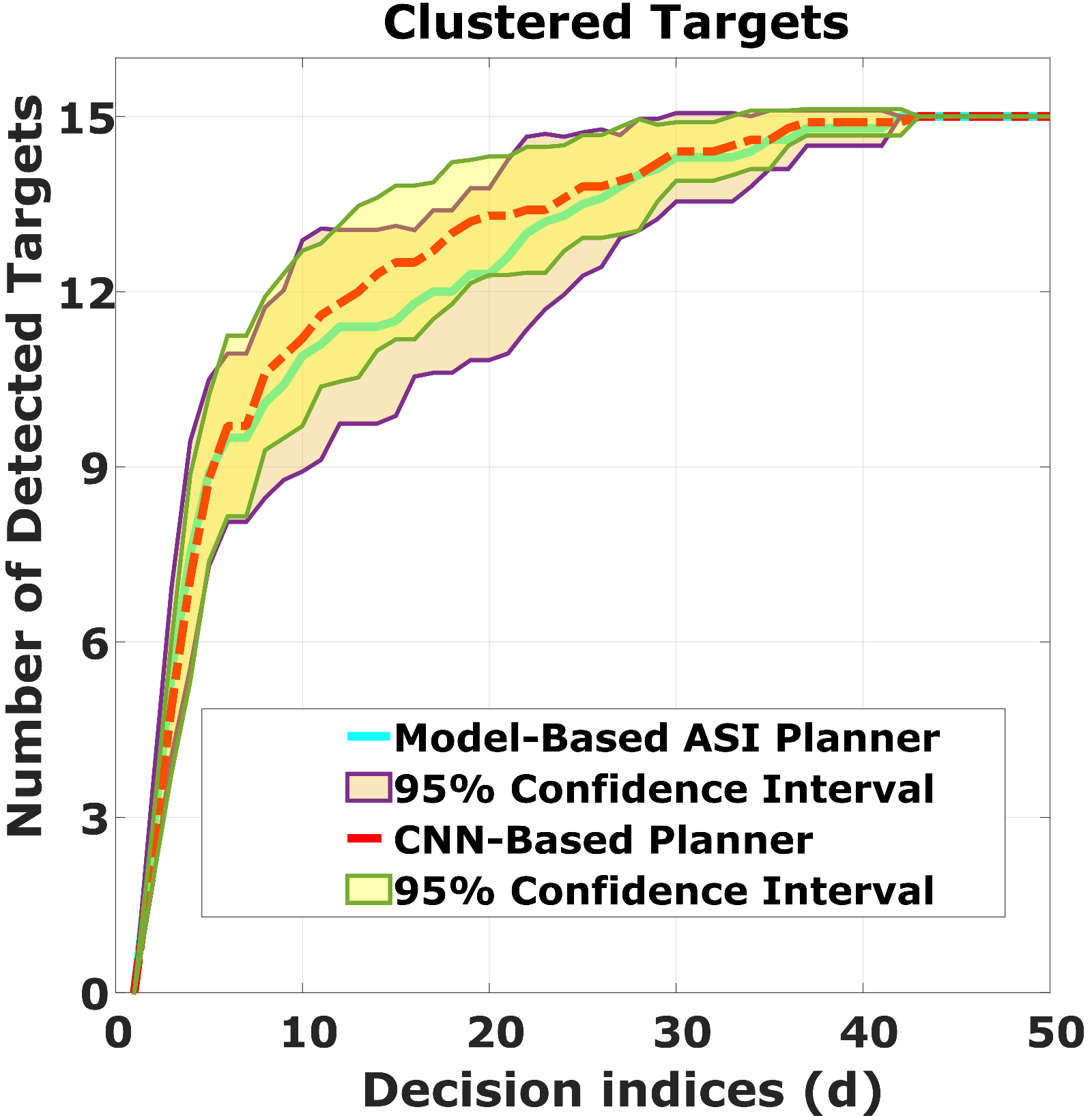}
    \caption{Number of targets detected using the CNN and ASI. Top: Uniformly random targets. Bottom: Clustered targets.}
    \label{FigureASI}
\end{figure}

\medskip \noindent \textit{E2: Comparison of CNN to ASI.} In this experiment, the CNN was trained on data generated by ASI. In ASI, the candidate set  $\mathcal{W}_d$ is defined as a $3 \times 3$ grid of potential waypoints, yielding nine discrete choices, spaced 80 meters apart along both axes and starting at $[10,10]^T$. The results in Fig. \ref{FigureASI} show the number of target detections as a function of the number of steps $k$. The CNN closely follows the behavior of the model-based planner. In terms of computation time, the ASI planner requires an average of $1.475\times 10^{-1}\pm 3.3 \times 10^{-2}$s per step, whereas the CNN exact time remains similar to the AS case, $1.38\times 10^{-4}\pm 1.04\times 10^{-4}$s. 

\begin{figure}[h!]
\centering
\includegraphics[scale=0.22]{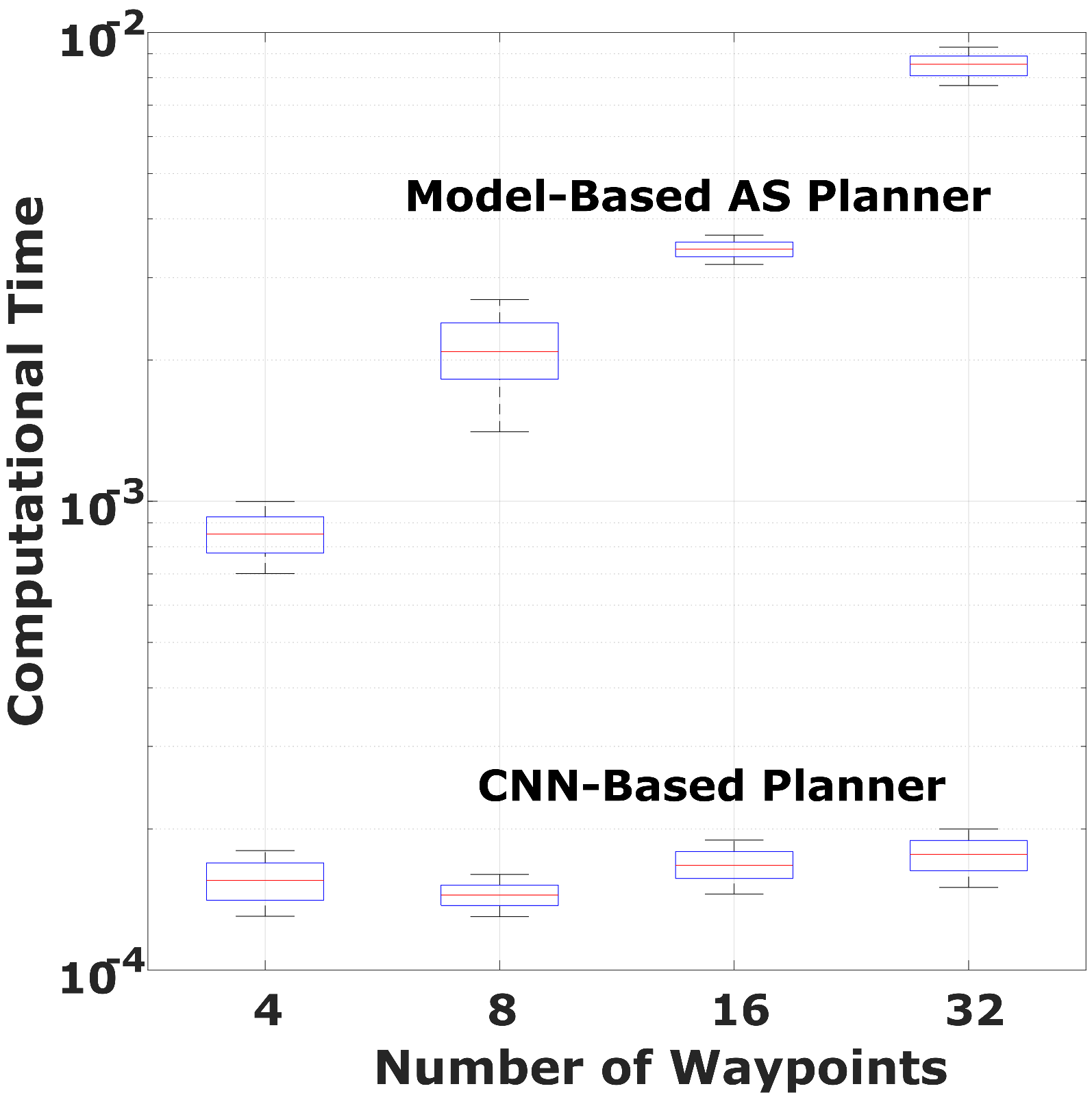}
\includegraphics[scale=0.22]{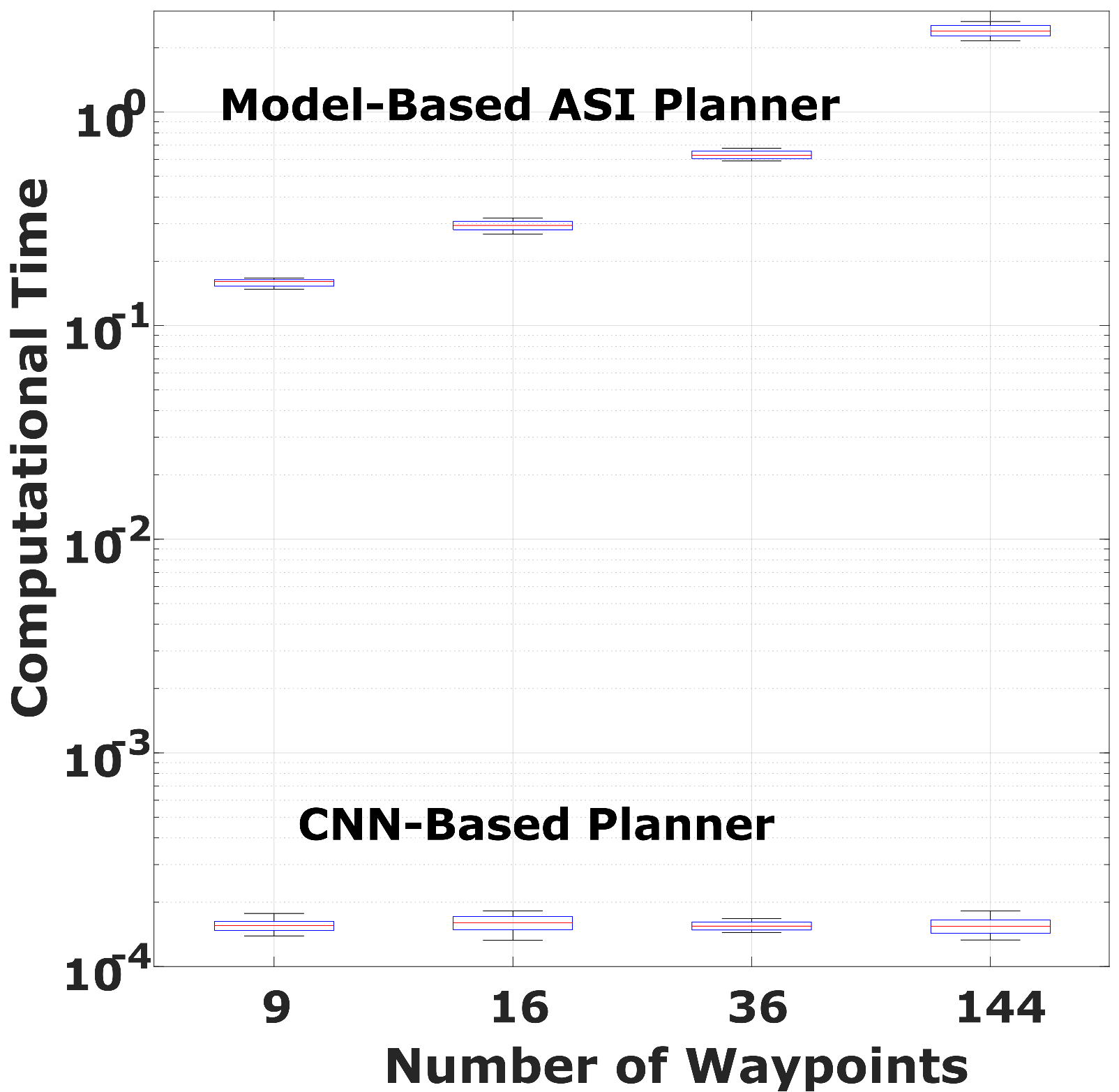}
\caption{Computational time per planning step as a function of the number of candidate waypoints. Top: AS vs.~CNN. Bottom: ASI vs.~CNN. Results are shown as boxplots over multiple trials (each trial contributes one averaged value across planning steps).}
\label{Figurecomputa}
\end{figure}

\medskip \noindent \textit{E3: Effect of Increasing the Number of Candidate Waypoints.} In this experiment, we investigate how increasing the number of candidate waypoints impacts the performance and computational cost of model-based planners AS and ASI, compared to the CNN. To ensure a fair comparison, the CNN was retrained for each candidate-set configuration using training data generated by running AS or ASI with that specific configuration. The evaluation is performed under the clustered target distribution, as explained before. For computational cost (Fig.~\ref{Figurecomputa}), results are based on a single run of 250 steps for AS and 50 steps for ASI, from which the mean computational time per planning step was computed.

For the AS planner, the number of candidate directions was varied across $4$, $8$, $16$, and $32$ choices—corresponding to angular separations of $90^\circ$, $45^\circ$, $22.5^\circ$, and $11.25^\circ$, respectively. For the ASI planner, the number of candidate waypoints was controlled by adjusting the waypoint spacing. Specifically, a spacing of $20$ m corresponds to $144$ candidate waypoints, $40$ m to $36$, $60$ m to $16$, and $80$ m to $9$ waypoints. In both planners, increasing the number of candidate waypoints improves detection performance since the planner generates finer motion, but it also increases computational cost.

The results in Fig.~\ref{Figurecomputa} illustrate how the number of candidate waypoints affects computational time for AS, ASI, and CNN. In both AS (top) and ASI (bottom), computational time rises sharply with candidate set size due to repeated online optimization, as reflected in the spread of the boxplots. In contrast, the CNN planner is unaffected by candidate set size and maintains a constant inference cost, since it outputs waypoints directly without enumeration. For training, the AS-based dataset contained $9{,}120$ samples and required about 12 minutes, while the ASI-based dataset contained $3{,}120$ samples and required about 5 minutes. Importantly, increasing the number of candidates in AS or ASI does not raise training cost, which depends only on the number of training samples rather than the candidate set size.

\begin{figure}[h!]
\centering
\includegraphics[scale=0.22]{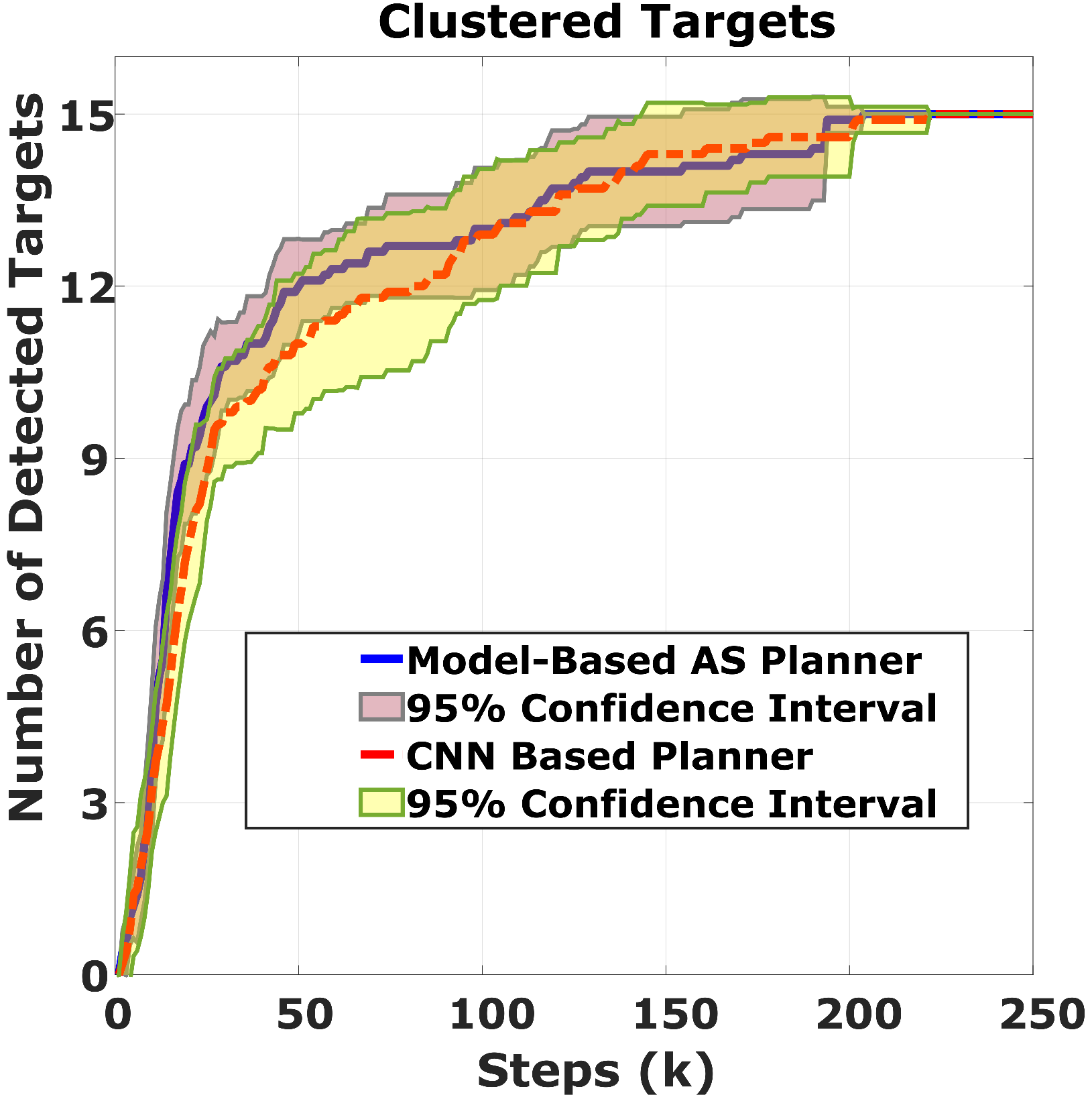}
\caption{Number of targets detected in clustered distribution using the CNN and AS (with 16 candidate choices).}
\label{Figureplanners}
\end{figure}

To illustrate performance for a longer number of waypoints, Fig.~\ref{Figureplanners} compares the CNN to AS with 16 candidate directions under a clustered target distribution. As expected, AS benefits from more candidate choices, achieving faster detection than with fewer candidates, see again Fig.~\ref{Figureplanner} in $E1$. The CNN-based planner matches this improved performance of AS.

\medskip \noindent \textit{E4: Ablation Study of CNN Input Channels.} In this experiment, we assess the contribution of each input channel by removing or simplifying it and evaluating the resulting detection performance. Recall the full CNN uses four input channels: exploration history, Gaussian-smoothed particle density, agent position, and boundary proximity. Since removing the particle density channel entirely would trivially degrade performance, we instead remove only the Gaussian smoothing while retaining the raw particle distribution. This isolates the contribution of smoothing itself. As for $E3$, $E4$ is also performed only for clustered targets.  Table~\ref{tab:ablation} summarizes the average number of targets detected together with 95\% confidence intervals.

\begin{table}[h!]
\centering
\label{tab:ablation}
\begin{tabular}{| m{3.25cm} | m{4.25cm}| }
\hline
\textbf{Ablation} & \textbf{Targets detected} \\
\hline
Original & 15 $\pm 0.2$ \\
\hline
No visitation history & 7.67  $\pm 1.45$ \\
\hline
No Gaussian Smoothing of intensity    & 8.87  $\pm 2.03$  \\
\hline
No agent position      & 10.87 $\pm 1.67$ \\
\hline
No boundary mask       & 12.09  $\pm 1.20$ \\
\hline
\end{tabular}
\caption{Target detection performance under CNN channel ablations.}
\end{table}

Among the four channels, the visitation history proves most critical, followed by the Gaussian smoothing. The removal of the agent position channel does not severely affect performance, likely because position information is implicitly encoded in the visitation history. Similarly, eliminating the boundary channel causes the agent to drift toward the edges, leaving several targets undetected.

\section{Conclusion} \label{sec:conclusion}
This paper introduced a CNN-based representation for multi-target search policies that reduces the computational complexity of existing model-based methods, enabling their deployment on resource-constrained hardware. Trained on model-based demonstrations, the CNN replaces online optimization with a direct state-to-waypoint mapping. Simulation results confirmed that the CNN matches the accuracy of the original model-based methods for both uniform and clustered target distributions, possibly after a brief regime where the performance slightly lags behind. The CNN approach is computationally-insensitive to the number of candidate waypoints. An ablation study highlighted the importance of our multi-channel CNN input design.

A limitation of the proposed approach is that the learned policy is specific to the task and dependent on the data distribution, thus likely requiring retraining when applied to new scenarios.
Future work will extend the framework to cooperative multi-agent settings and validate performance in real-life experiments, e.g.~in the setup used for our model-based approach in \cite{Bilal:25}. It would also be interesting to analyze the error introduced by the CNN approximation.


\appendix
\section{CNN Architecture and Training Details}
\label{sec:appendix}

This appendix summarizes the complete CNN architecture and training configuration used in the experiments. The network takes as input a tensor of size $26 \times 26 \times 4$, corresponding to the four channels described in Section~3: visitation history, Gaussian-smoothed particle density, agent position, and boundary proximity. The output is a 2-dimensional vector in $[0,1]^2$ representing the normalized waypoint coordinates.

Table~\ref{tab:cnn_architecture} provides the layer-by-layer specification of the CNN used in Fig.~\ref{neural}. 

\begin{table}[h]
\centering
\caption{Detailed CNN architecture used for waypoint prediction.}
\label{tab:cnn_architecture}
\begin{tabular}{| m{1.6cm} | m{1.85cm} | m{1.85cm}|}
\hline
Layer & Parameters  & Output size \\
\hline
Input & --- &  $26 \times 26 \times 4$ \\
\hline
Conv1 & 32 filters, $5 \times 5$, stride 1, \textit{same} padding &  $26 \times 26 \times 32$ \\
\hline
BatchNorm1 & --- &  $26 \times 26 \times 32$ \\
\hline
LeakyReLU1 & slope $0.01$ &  $26 \times 26 \times 32$ \\
\hline
MaxPool1 & $2 \times 2$, stride 2 &  $13 \times 13 \times 32$ \\
\hline
Conv2 & 64 filters, $3 \times 3$, stride 1, same padding &  $13 \times 13 \times 64$ \\
\hline
BatchNorm2 & --- &  $13 \times 13 \times 64$ \\
\hline
LeakyReLU2 & slope $0.01$ &  $13 \times 13 \times 64$ \\
\hline
Dropout & probability $0.5$ &  $13 \times 13 \times 64$ \\
\hline
Flatten & --- &  $10816$ \\
\hline
Fully connected 1 & 128 units &  $128$ \\
\hline
LeakyReLU3 & slope $0.01$ &  $128$ \\
\hline
Fully connected 2 & 2 units &  $2$ \\
\hline
Sigmoid output & maps output to $[0,1]^2$ &  $2$ \\
\hline
\end{tabular}
\end{table}

This architecture was selected as a compromise between representational capacity and low inference cost. In particular, the two convolutional layers extract local spatial patterns from the belief map and visitation structure, while the fully connected layers map these features to continuous waypoint coordinates. The use of batch normalization improves optimization stability, and dropout reduces overfitting.

Training to optimize \eqref{opt} was performed using the Adam optimizer \citep{KingmaBa2015}. The optimizer hyperparameters were:
\begin{itemize}
    \item initial learning rate: $10^{-3}$
    \item learning-rate schedule type: piecewise decay
    \item decay period:  every $30$ epochs
    \item decay factor: $0.1$
    \item mini-batch size: $64$
    \item maximum number of epochs: $100$
\end{itemize}
To prevent overfitting, early stopping was employed based on the validation loss. A validation split of $10$--$20\%$ of the training data was used, and training was terminated if the validation loss did not improve by at least $10^{-4}$ for $5$--$10$ consecutive epochs (patience).

\balance
\makeatletter
\makeatother
{\small\bibliography{ifacconf}}            
                                                                        
\end{document}